\pdfoutput=1

\documentclass[11pt]{article}

\usepackage[final]{acl}

\usepackage{times}
\usepackage{latexsym}
\usepackage{amsmath}
\usepackage{amssymb}
\usepackage{bm}

\usepackage[T1]{fontenc}

\usepackage[utf8]{inputenc}

\usepackage{microtype}

\usepackage{inconsolata}
\usepackage{tcolorbox}
\usepackage{booktabs}

\usepackage{graphicx}
\usepackage{subcaption}
\usepackage{caption}

\title{Whispering Context: Distilling Syntax and Semantics for Long Speech Transcripts}

\author{Duygu Altinok \\
  Independent Researcher, Germany \\
  \texttt{duygu.altinok@onlyduygu.com}}

\begin{document}
\maketitle
\begin{abstract}
ASR systems often struggle with maintaining syntactic and semantic accuracy in long audio transcripts, impacting tasks like Named Entity Recognition (NER), capitalization, and punctuation. We propose a novel approach that enhances ASR by distilling contextual knowledge from LLaMA models into Whisper. Our method uses two strategies: (1) token-level distillation with optimal transport to align dimensions and sequence lengths, and (2) representation loss minimization between sentence embeddings of Whisper and LLaMA, blending syntax and semantics. Evaluations on the Spoken Wikipedia dataset—a benchmark with long audios and rich entities—demonstrate significant improvements in Word Error Rate (WER), NER, capitalization, and punctuation success. By introducing novel NER metrics and exploring semantics-aware ASR, our work highlights the value of integrating linguistic context into transcription, setting a foundation for robust, context-aware ASR in long-form speech.
\end{abstract}
%
Large Language Models (LLMs) have demonstrated remarkable performance in tasks such as question answering and summarization, often outperforming specialized models \cite{openai2024gpt4technicalreport,touvron2023llamaopenefficientfoundation,brown2020languagemodelsfewshotlearners}. Their extensive linguistic knowledge, derived from vast text corpora, has motivated efforts to adapt LLMs for speech-related tasks, particularly in enhancing Automatic Speech Recognition (ASR).

End-to-End (E2E) ASR models \cite{yadav2020endtoendnamedentityrecognition,Gaido_2023,koluguri2024longernotnecessarilystronger} aim to produce fully formatted transcriptions, including punctuation, capitalization, and proper number formatting, aligning with user expectations and supporting downstream NLP tasks like machine translation and summarization. Unlike conventional systems reliant on text-only post-processing, E2E models can leverage acoustic cues (e.g., prosody) to improve formatting, reducing errors in complex cases such as numerical or entity-specific formatting. For instance, “five five five one two two four” may be formatted as “555-1224” for a phone number but as “5551224” for a student ID. This highlights the dual challenge of recognizing entities and applying appropriate formats, underscoring the complexity of fully formatted ASR transcriptions.

However, developing such models requires large-scale, high-quality paired data, which is often scarce. Datasets like the Spoken Wikipedia \cite{spokenwiki}, containing 395 hours of professionally transcribed audio with diverse entity types, remain insufficient for modern ASR research. Whisper \cite{radford2022robustspeechrecognitionlargescale}, trained on 680,000 hours of audio, achieves strong transcription performance but lacks text normalization and entity formatting. Even with its scale, Whisper remains limited compared to text-based LLMs.

To bridge this gap, we propose distilling the LLaMA family’s \cite{touvron2023llamaopenefficientfoundation} extensive text knowledge into Whisper, enabling it to produce fully formatted transcriptions for long-form audio. Our approach involves dividing audio and transcripts into 30-second chunks to align with Whisper’s training framework. We enrich the training data with entity annotations (e.g., \texttt{<LOC>United States</LOC>}), tagging 20 entity types using spaCy’s \cite{spacy2} Named Entity Recognition (NER) model trained on OntoNotes \cite{ontonotes}, supplemented with pattern-based extractors for email and URL entities. This enriched data allows Whisper to learn both entity recognition and formatting. 

\begin{figure*}[t]
  \centering
  \includegraphics[width=\linewidth]{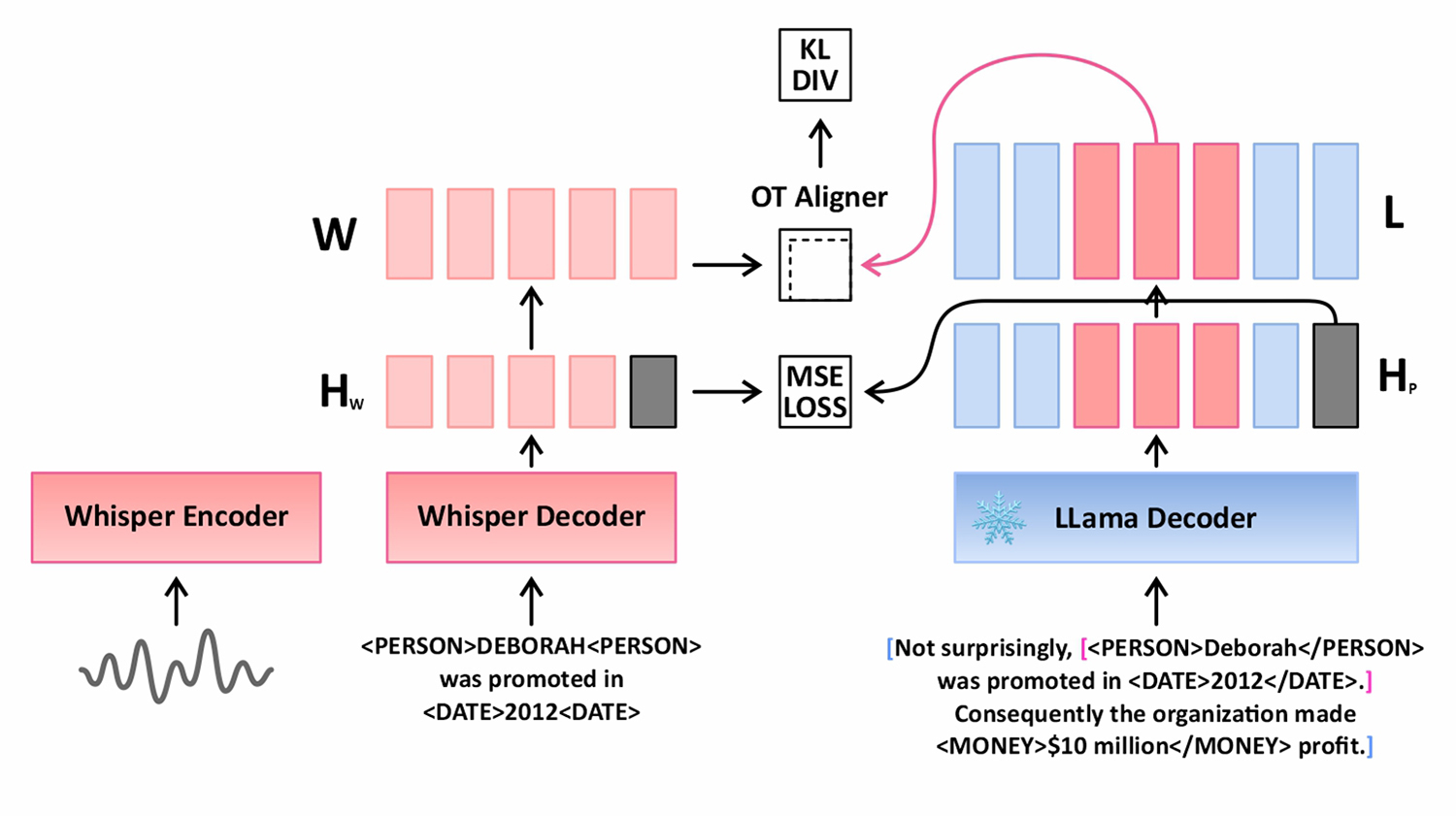}
  \caption{The architecture of the proposed model.}
  \label{fig:speech}
\end{figure*}

Our proposed distillation strategy integrates LLaMA’s syntactic and semantic capabilities into Whisper using two complementary methods:
\begin{itemize}
  \item  Aligning LLaMA token logits with Whisper token logits using Kullback-Leibler (KL) divergence. This process incorporates an optimal transport algorithm \cite{villani2008optimal} to address modality differences between text and speech.
  \item  Minimizing Mean Squared Error (MSE) loss between LLaMA’s context embeddings and Whisper’s. We extend LLaMA’s embeddings to include preceding and succeeding text tokens, experimenting with varying window sizes to explore the impact of broader context on transcription quality.
\end{itemize}

A key innovation of our approach is the use of an extended context window during representation-level distillation to provide Whisper with richer textual clues. Instead of distilling knowledge solely from LLaMA’s representation of the chunk text (e.g., a 30-second segment), we incorporate a broader context from the entire transcript. Specifically, for a fixed context size of tokens, we include half the tokens preceding the chunk and half following it within the transcript, effectively expanding the semantic scope available during distillation. We experimented with various context sizes—64, 128, 256, 512, and 1024 tokens—and observed that increasing the context size significantly improved NER performance by enabling Whisper to leverage more semantic information from the surrounding text. This extended context trick proved particularly effective for handling long-tail entities, where global transcript-level understanding is critical for accurate recognition and formatting. Figure \ref{fig:speech} exhibits an overview of our approach.

To evaluate our approach, we assess both syntactic (e.g., punctuation and capitalization) and semantic (e.g., entity recognition and formatting) success. Numerical entities are evaluated using Character Error Rate (CER), while worded entities (e.g., locations) are assessed using Jaro-Winkler similarity \cite{jaro1989, winkler1990}. SeqEval metrics \cite{seqeval} measure entity recognition performance, while sentence-level syntax tasks evaluate punctuation and capitalization.

Our contributions can be summarized as follows:
\begin{itemize}
\item We propose a robust distillation framework that transfers LLaMA’s linguistic knowledge into Whisper, aligning sequence lengths and embedding dimensions using optimal transport.
\item We leverage an extended context window during representation-level distillation, incorporating tokens from both before and after the chunk text in the transcript. This approach provides richer semantic clues, significantly improving NER performance, particularly for long-tail entities.
\item  We introduce the novel task of entity formatting, combining entity recognition with type-specific formatting, supported by detailed evaluation metrics.
\item  We assess numerical and non-numerical entities separately, using metrics tailored to their specific characteristics.
\item We demonstrate how integrating syntactic and semantic knowledge from LLMs improves ASR transcription quality for long-form audio.
\end{itemize}

Our results reveal significant improvements across all evaluated metrics, demonstrating how LLMs can enhance E2E ASR performance by combining semantic understanding and syntactic accuracy. To our knowledge, this is the first study to use LLMs for enhancing both sentence syntax and entity semantics in ASR, introducing novel metrics and advancing the state of context-aware transcription.

\section{Dataset and Data Preparation}
\begin{figure*}[t]
  \centering
  \includegraphics[width=0.6\linewidth]{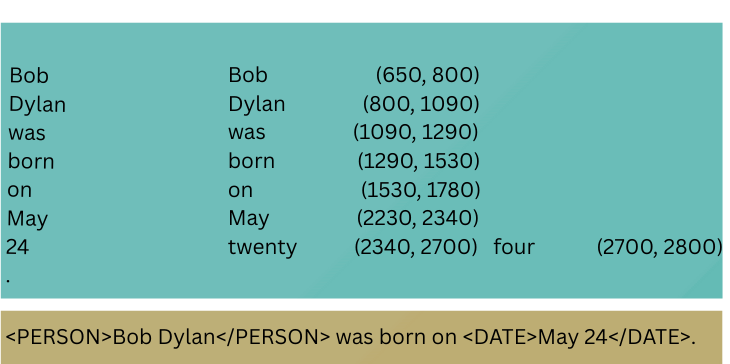}
  \caption{An example sentence from the dataset.}
  \label{fig:instance}
\end{figure*}

This study utilizes the Spoken Wikipedia dataset, chosen for its long-form audio, rich
entity vocabulary, and diverse entity types. The dataset comprises 1,339 English Wikipedia
articles (395 hours of audio) and provides aligned audio and text in both normalized and
formatted forms. It includes punctuation as separate tokens and word-level timestamps
in XML format. An example instance from the dataset is illustrated in Figure \ref{fig:instance}.

On average, each instance contains 18 minutes of audio and 2,613 words, which
aligns with the typical structure of Wikipedia articles. This characteristic provides sub-
stantial contextual information, making the dataset particularly suitable for semantic
analysis.

We partitioned the dataset into 1,000 training instances (333 hours) and 241 test in-
stances (62 hours). The corpus contains 3.5M words, including 472K entities spanning
828K words, accounting for 23\% of the total word count. Common entity types include
person (103K), date (71K), location (60K), organization (59K), and cardinal number
(42K). Numerical entity types collectively account for approximately 80K entities, rep-
resenting 16\% of all entity types.

Using formatted transcripts, we implemented an entity formatter with a NER module and tag embedder to label transcripts as exhibited in Figure \ref{fig:instance}. To experiment with different context window sizes for LLaMA, we extracted surrounding text for each chunk.

A total of 22 NER labels were used: 18 derived from spaCy NER and 4 additional custom labels (URL, email, phone number, and generic numeric type). Entities were tagged using spaCy NER, while custom types were annotated using a regex-based extractor.

The dataset is particularly rich in numerical entity formats, which include examples
such as \textit{15,981.21}, \textit{3,000}, \textit{9.2\%}, \textit{2.1 kg}, and \textit{555-1142}, as well as email addresses and URLs. These types of entities are uncommon in standard speech datasets, making this corpus uniquely suited for tasks involving numerical and specialized entity recognition.

\section{Proposed Method}
\subsection{Overview}
We distill task-specific knowledge from three teacher models (LLaMA 1B, 3B, and 8B) into the medium-sized Whisper model using three strategies: token-level distillation \cite{gou2021knowledge}, representation-level distillation \cite{gou2021knowledge}, and a hybrid approach. The process involves: (i) fine-tuning the teacher models on formatted transcripts to adapt them to the task, (ii) fine-tuning the student on transcript-audio pairs to introduce task-specific knowledge, (iii) freezing teacher layers during student training to preserve task-specific knowledge, (iv) aligning Whisper’s token logits and hidden representations with LLaMA’s, and (v) training the student using a combination of classification and distillation losses. During steps (i) and (ii), the tokenizer’s vocabulary is augmented with entity tags to enhance entity recognition and formatting. Further details on fine-tuning and freezing strategies are provided in the experiments section.

\subsection{Aligning LLaMa representations with Whisper}
To align Whisper and LLaMA logits for distillation, we start with an input audio-transcript pair $(\mathbf{X}, \mathbf{T})$, where $\mathbf{X}$ represents the input speech waveform and $\mathbf{T} = \{t_1, t_2, \dots, t_{T_t}\}$ represents the corresponding transcript tokens of length $T_t$. Whisper processes the audio $\mathbf{X}$ and generates hidden states (embeddings), represented as $\mathbf{H}_w = \{h_{w,1}, h_{w,2}, \dots, h_{w,T_w}\} \in \mathbb{R}^{T_w \times d_w}$, where $T_w$ is the sequence length of the speech embeddings and $d_w$ is the embedding dimension. Similarly, LLaMA processes the transcript $\mathbf{T}$ and generates hidden states $\mathbf{H}_l = \{h_{l,1}, h_{l,2}, \dots, h_{l,T_t}\} \in \mathbb{R}^{T_t \times d_l}$.

The embeddings are projected to logits via linear transformations:
\begin{equation}
\mathbf{W} = \mathbf{H}_w \cdot \mathbf{U}_w, \quad \mathbf{L} = \mathbf{H}_l \cdot \mathbf{U}_l,
\end{equation}
where $\mathbf{W} \in \mathbb{R}^{T_w \times V_w}$ and $\mathbf{L} \in \mathbb{R}^{T_t \times V_l}$ are the logits for Whisper and LLaMA, and $\mathbf{U}_w \in \mathbb{R}^{d_w \times V_w}$ and $\mathbf{U}_l \in \mathbb{R}^{d_l \times V_l}$ are the vocabulary projection matrices for Whisper and LLaMA, respectively.

To overcome the mismatch in vocabulary sizes ($V_w$ and $V_l$) and sequence lengths ($T_w$ and $T_t$), we project the logits into a shared $d$-dimensional embedding space:
\begin{equation}
\mathbf{W}' = \mathbf{W} \cdot \mathbf{P}_w, \quad \mathbf{L}' = \mathbf{L} \cdot \mathbf{P}_l,
\end{equation}
where $\mathbf{P}_w \in \mathbb{R}^{V_w \times d}$ and $\mathbf{P}_l \in \mathbb{R}^{V_l \times d}$ are trainable projection matrices.

After projection, we compute the Optimal Transport (OT) cost between $\mathbf{W}'$ and $\mathbf{L}'$ using the Sinkhorn algorithm \cite{sinkhorn1967concerning}:
\begin{equation}
\mathcal{L}_{\text{OT}} = \text{WassersteinDistance}(\mathbf{W}', \mathbf{L}'),
\end{equation}
where $\mathcal{L}_{\text{OT}}$ quantifies the alignment cost between the two distributions, ensuring that Whisper and LLaMA logits are mapped into the same shared space for effective distillation. As part of this process, we also compute the optimal alignment, extract the aligned $\mathbf{L}'$ from the shared space, and project it back to Whisper's dimension. This yields the final aligned logits $\mathbf{L}''$ with shape $(T_w, V_w)$.

\subsection{Distilling the knowledge of LLaMA}
For distillation we used two strategies, token-level distillation and representation-level distillation. We give details of how the losses are calculated for both.

\noindent\textbf{Token-level loss} In this type of distillation, the student is trained to produce logits that mimic the teacher’s predictions. To increase the entropy of model outputs and extract richer information, we apply a temperature scaling to the logits and minimize the Kullback-Leibler (KL) divergence between the teacher and student outputs.
Hence after obtaining aligned LLaMa logits $\mathbf{L}''$ , we applied temperature, softmaxed then we calculated Kullback-Liebler divergence between these logits and Whisper logits. More formally:
\begin{equation}
  \label{eqn:softmax}
  p_{L''}(t) = \text{softmax}\left(\frac{\mathbf{L}''_t}{T}\right), \quad p_{W}'(t) = \text{softmax}\left(\frac{\mathbf{W}'}{T}\right).
\end{equation}
The Kullback-Leibler (KL) divergence at time step $t$ is defined as:
\begin{equation}
D_{\text{KL}}(p_{L''}(t) \, \Vert \, p_{W'}(t)) = \sum_{i=1}^{V_w} p_{L''}(t)_i \log \frac{p_{L''}(t)_i}{p_{W'}(t)_i},
\end{equation}
where $p_{L''}(t)_i$ and $p_W'(t)_i$ are the probabilities assigned to the $i$-th token in Whisper's vocabulary by the teacher and student, respectively. The total KL divergence loss $\mathcal{L}_{\text{KL}}$ is computed by summing over all time steps and scaling by the temperature factor:
\begin{equation}
\mathcal{L}_{\text{KL}} = \sum_{t=1}^{T_w} D_{\text{KL}}(p_{L''}(t) \, \Vert \, p_{W'}(t)),
\end{equation}
where the factor $T^2$ compensates for the scaling introduced by the temperature during softmax in Equation \ref{eqn:softmax}.

\noindent\textbf{Representation loss} This approach aligns the intermediate representations of the teacher and student. Here we take sentence embedding as the final element of the hidden states, denote them as $h_{W}$ and $h_{L}$ respectively. Then a learnable projection matrix is applied to teacher embeddings to overcome the dimension mismatch.  Then we calculate MSE loss between these two embeddings:
\begin{equation}
\mathcal{L}_{\text{rep}} = \lVert \mathbf{h}^{(W)} - \mathbf{W}_{\text{proj}} \mathbf{h}^{(L)} \rVert_2^2 .
\end{equation}
\textbf{Final loss. } Final loss is a sum of two distillation losses, optimal transport loss, and label loss of Whisper. 
\begin{equation}
\label{eqn:loss}
\begin{aligned}
\mathcal{L}_{\text{CE}} &= -\frac{1}{N} \sum_{i=1}^N \sum_{j=1}^C y_{ij} \log \hat{y}_{ij}, \\
\mathcal{L}_{\text{total}} &= \alpha \mathcal{L}_{\text{KL}} + \beta \mathcal{L}_{\text{rep}} + \gamma \mathcal{L}_{\text{OT}} + (1-\alpha-\beta-\gamma) \mathcal{L}_{\text{CE}},
\end{aligned}
\end{equation}

\subsection{Evaluation Metrics}
We use SeqEval to evaluate NER, F1 score and accuracy for capitalization and punctuation, CER for numerical entities, and Jaro-Winkler similarity for non-numerical entities to assess transcription quality across syntax and semantics.
Entity formatting is evaluated by type: numerical entities use Character Error Rate (CER) for strict digit-level precision, while non-numerical entities use Jaro-Winkler similarity to measure string similarity and tolerate minor variations. This ensures accurate assessment tailored to each entity type.

\subsection{Experimental Setup}

\noindent \textbf{Fine-Tuning LLaMA Models} We fine-tuned LLaMA teacher models (1B, 3B, and 8B) using parameter-efficient techniques and 4-bit quantization via BitsandBytes\footnote{\url{https://huggingface.co/docs/transformers/main/quantization/bitsandbytes}}. This included double quantization in the NF4 format \cite{dettmers2023qloraefficientfinetuningquantized} with bfloat16 computations for stability. Using HuggingFace's PEFT library \footnote{\url{https://huggingface.co/docs/peft/index}}, we applied LoRA \cite{hu2021loralowrankadaptationlarge} focusing on attention layers with rank-16 matrices, a scaling factor of 32, and a dropout rate of 0.01. Training was conducted for one epoch using AdamW (learning rate $2e-4$, weight decay 0.01). LLaMA tokenizers were augmented with 44 entity tags across 22 entity types, and transcripts were processed in 512-token chunks with a stride of 20 to preserve context.

\noindent \textbf{Fine-Tuning Whisper} The medium-sized Whisper model was fine-tuned using HuggingFace Trainer with a batch size of 32, learning rate $1e-4$, 1250 warmup steps, and 100 epochs. Entity tags were added to the Whisper tokenizer. Transcripts were divided into 30-second chunks paired with corresponding audio. This model is referred to as Whisper-tuned.

\noindent \textbf{Distillation} We distilled LLaMA knowledge into Whisper-tuned using a single training epoch, the AdamW optimizer (learning rate $2e-5$, weight decay 0.01), and a batch size of 32. The distillation process optimized the loss defined in Equation \ref{eqn:loss}, with coefficients $\alpha$, $\beta$, and $\gamma$ controlling the weighting of token-level, representation-level, and classification losses. We experimented with different values of these coefficients to identify the optimal values for balancing distillation components. Additionally, we experimented with various context lengths (64, 128, 256, 512, and 1024 tokens), where the chunk text was placed at the center of surrounding tokens from the full transcript. This model is referred to as Whisper-distilled.

\noindent \textbf{Inference} Whisper-distilled operates independently of LLaMA during inference. It functions as a standard HuggingFace Whisper model and uses the Hugging Face inference pipeline\footnote{\url{https://huggingface.co/docs/transformers/v4.43.4/en/main_classes/pipelines}}. Long audio inputs are processed in 30-second chunks with a 5-second stride.

\noindent \textbf{Hardware and Code} All experiments were conducted on a single NVIDIA H100 GPU. Code is available on GitHub \footnote{\url{https://github.com/DuyguA/ASRU2025-Whispering-Context}}, and fine-tuned models can be accessed on HuggingFace \footnote{\url{https://huggingface.co/collections/BayanDuygu/2025-asru-whispering-context-6807782eeedf66921e64864e}}.

\section{Results and Discussion}
\subsection{Baseline}
Whisper-Medium achieved a 0.38 WER on The Spoken Wiki dataset, which is reasonable given the complexity of the transcripts. Our fine-tuned version reduced the WER to 0.26. For WER calculations, we removed punctuation, NER tags, and converted text to lowercase.

Fine-tuning had minimal impact on punctuation and capitalization. Capitalization performance was already strong (Table \ref{tab:punct-results}), while punctuation prediction remained poor. Exclamation marks and semicolons, though present in Whisper's vocabulary, were never predicted. Note that punctuation marks within numbers or entities (e.g., commas in amounts) were excluded from evaluation. 

\begin{table*}[th]
  \caption{Comparison of punctuation statistics between Whisper-tuned and Whisper-distilled models. Reference counts are shared across both models. Whisper-Medium achieved the same performance as Whisper-tuned and is therefore omitted.}
  \label{tab:punct-results}
  \centering
  \setlength{\tabcolsep}{2pt}
  \begin{tabular}{l | r | r r r r | r r r r}
    \toprule
    \textbf{Punct. Mark} & \textbf{Ref. Cnt} & \multicolumn{4}{c|}{\textbf{Whisper-tuned}} & \multicolumn{4}{c}{\textbf{Whisper-distilled}} \\
    \cmidrule(lr){3-6} \cmidrule(lr){7-10} &  & \textbf{Acc.} & \textbf{F1} & \textbf{Prec.} & \textbf{Recall}  & \textbf{Acc.} & \textbf{F1} & \textbf{Prec.} & \textbf{Recall} \\
    \midrule
    Capitals    & 1593 & 0.98 & 0.99 & 1.00  & 0.97   & 0.99  & 0.99  & 1.00 & 0.99 \\
    Comma       & 6031 & 0.48 & 0.65 & 0.57  & 0.76   & 0.68  & 0.75  & 0.80 & 0.71 \\
    Period      & 4429 & 0.67 & 0.80 & 0.73  & 0.89   & 0.85  & 0.90  & 0.92 & 0.88 \\
    Semicolon   & 201  & 0.00 & 0.00 & 0.00  & 0.00   & 0.61  & 0.61  & 0.62 & 0.60 \\
    Excl. Mark  & 8    & 0.00 & 0.00 & 0.00  & 0.00   & 1.00  & 1.00  & 1.00 & 1.00 \\ 
    Que. Mark   & 18   & 0.41 & 0.58 & 0.69  & 0.50   & 0.96  & 0.95  & 0.96 & 0.94 \\
    \bottomrule
  \end{tabular}
\end{table*}

For NER, the base model did not emit labels, so results are only available for the Whisper-tuned version (Table \ref{tab:ner-results}). Numerical entities, such as cardinal numbers and quantities along with dates performed well due to clear wording and minimal context dependence. Surprisingly, person names were reasonably accurate, likely due to Wikipedia’s bias toward celebrity names. Person names are expected to perform worse due to variability in names and regional pronunciations. Also transcription errors, especially for rare or multi-word names, further reduce precision and recall. However this was not the case for this corpus due to genre. GPE entities (e.g., countries, cities) perform well due to their consistent patterns, frequent representation in training data, and clear context in speech.

\begin{table*}[th]
  \caption{Comparison of NER statistics between Whisper-tuned and Whisper-distilled models on the most common entity types of the corpus.}
  \label{tab:ner-results}
  \centering
  \setlength{\tabcolsep}{4pt}
  \begin{tabular}{l | r | r r r | r r r}
    \toprule
    \textbf{NER Label} & \textbf{Ref. Cnt} & \multicolumn{3}{c|}{\textbf{Whisper-tuned}} & \multicolumn{3}{c}{\textbf{Whisper-distilled}} \\
    \cmidrule(lr){3-5} \cmidrule(lr){6-8}
    &  & \textbf{F1} & \textbf{Prec.} & \textbf{Recall} & \textbf{F1} & \textbf{Prec.} & \textbf{Recall} \\
    \midrule
    PERSON   & 4212  & 0.87 & 0.88 & 0.85 & 0.95 & 0.96 & 0.93  \\
    DATE     & 3383  & 0.85 & 0.87 & 0.84 & 0.94 & 0.95 & 0.93  \\
    GPE      & 2462  & 0.85 & 0.87 & 0.80 & 0.92 & 0.93 & 0.90  \\
    CARDINAL & 2212  & 0.88 & 0.89 & 0.87 & 0.99 & 0.99 & 0.99  \\
    QUANTITY & 351   & 0.91 & 0.88 & 0.94 & 0.99 & 0.99 & 0.99  \\
    MONEY    & 109   & 0.82 & 0.83 & 0.81 & 0.94 & 0.94 & 0.94  \\
    \midrule
    FAC      & 405   & 0.65 & 0.68 & 0.62 & 0.80 & 0.82 & 0.78 \\  
    PRODUCT  & 386   & 0.70 & 0.72 & 0.68 & 0.85 & 0.87 & 0.83 \\
    LAW     &  135   & 0.55 & 0.58 & 0.52 & 0.70 & 0.73 & 0.68 \\
    \bottomrule
  \end{tabular}
\end{table*}

Looking at the entity formatting success (Table \ref{tab:jw_stats}), CER values for numerical entities were higher than expected. Errors were often due to incorrect punctuation in numbers (e.g., "1.2.211 million") or long-worded numbers (e.g., monetary amounts) split across chunk boundaries. While Whisper performs well on formatted numbers, there is still room for improvement. Coming to the textual entities, we see that dates are quite successful, due to easy wording, month, day names with numbers. Being religious, ethical and political groups (i.e. American, Democrat, Muslim) NORP is recognized well due to not very variable wordings too. However person names, organization names, event and location names performed suboptimal due to long and complex names (International Federation of Red Cross and Red Crescent Societies),  variability of wordings and also including some long-tail words (Chattanooga, a city in Tenessee, USA). 

\begin{table*}[th]
  \caption{Comparison of Jaro-Winkler (JW) and CER scores for two models on the most common entity types of the corpus.}
  \label{tab:jw_stats}
  \centering
  \begin{tabular}{l | r r | l | r r}
    \toprule
    \textbf{Label (JW)} & \textbf{Whisper-tuned} & \textbf{Whisper-distill} & \textbf{Label (CER)} & \textbf{Whisper-tuned} & \textbf{Whisper-distill} \\
    \midrule
    PERSON         & 0.75 & 0.81 & CARDINAL       & 0.25 & 0.02 \\
    DATE           & 0.94 & 0.96 & NUMERIC        & 0.22 & 0.05 \\
    GPE            & 0.85 & 0.88 & TIME           & 0.41 & 0.08 \\
    NORP           & 0.95 & 0.97 & QUANTITY       & 0.22 & 0.09 \\ 
    ORG            & 0.71 & 0.76 & MONEY          & 0.42 & 0.09 \\
    EVENT          & 0.75 & 0.81 & PERCENT        & 0.32 & 0.06 \\
    LOC            & 0.67 & 0.77 & URL            & 0.57 & 0.23 \\
    \midrule
    FAC            & 0.70 & 0.80 & & & \\
    PRODUCT        & 0.65 & 0.85 & & & \\
    LAW            & 0.60 & 0.75 & & & \\
    \bottomrule
  \end{tabular}
\end{table*}

\section{Distillation results}
Here we present the results of our approach. First we discuss the effects of distillation hyperparameters, then dive into effect of transcript context length and teacher model size.

\subsection{Effect of distillation hyperparameters}
We evaluated the distillation process by fixing the optimal transport loss multiplier ($\gamma$) at a small value (0.001) while varying the coefficients for logits loss ($\alpha$) and representation loss ($\beta$). To determine the best $\gamma$, we conducted a hyperparameter search with fixed $\alpha$ and $\beta$ values and found that $\gamma$ should remain minimal, below 0.005.

For small $\alpha$ and $\beta$ values, the student model achieved significant improvements, with a WER/NER F1 of 0.20/0.82 compared to the baseline of 0.26/0.61. The optimal values were $\alpha = 0.005$ and $\beta = 0.005$, demonstrating the effectiveness of auxiliary signals from the teacher’s logits and representations. However, increasing $\alpha$ or $\beta$ beyond small values led to sharp performance drops, likely due to overemphasis on auxiliary losses interfering with the primary task.

\subsection{Semantic Success}
We previously noted that the WER dropped to 0.20, but the most significant gains were observed in punctuation, as shown in Table \ref{tab:punct-results}.

For commas and periods, Whisper-distilled outperforms Whisper-tuned across all metrics, demonstrating improved handling of sentence and clause boundaries. Commas, which rely heavily on prosody (e.g., pauses, intonation) and semantic context, show the most substantial improvement. For periods, Whisper-tuned achieves an F1 score of 0.80, while Whisper-distilled improves this to 0.90, likely due to better semantic reasoning and the ability to recognize sentence-ending cues.

Semicolons, which depend more on syntactic cues than acoustic features, are poorly handled by Whisper-tuned (F1: 0.00). Whisper-distilled, however, achieves an F1 score of 0.61, highlighting the impact of LLaMA’s syntactic understanding. For exclamation marks and question marks, Whisper-tuned struggles (F1: 0.00 and 0.58, respectively), as acoustic cues alone are insufficient. Whisper-distilled achieves perfect performance for exclamation marks (F1: 1.00) and near-perfect results for question marks (F1: 0.95), showcasing the importance of contextual reasoning for these punctuation marks.

Both models perform well on capitalization, indicating that this task relies more on acoustic and textual cues, such as sentence boundaries, rather than deep semantic reasoning.

NER also improved significantly, as seen in Table \ref{tab:ner-results}. Numerical entities showed dramatic gains, with near-zero CER across all types (Table \ref{tab:jw_stats}). Testing confirmed that most numerical entities were formatted correctly, including punctuation within entities (e.g., "4-11 May", "\$5.2 million", "https://www.apache.org/licenses"), with minimal insertions or deletions. These results suggest that LLaMA enhanced Whisper’s semantic understanding of entity structure.

For textual entities, all types improved, with substantial gains for locations (LOC) and events. Locations, such as cities and landmarks, require both phonetic accuracy and semantic context (e.g., "in Paris" or "Mount Everest"), which the distilled model leveraged to reduce transcription errors. Events, often represented by multi-word phrases (e.g., "World Cup Final"), also benefited from the contextual reasoning introduced by LLaMA-distilled Whisper. The use of Jaro-Winkler similarity highlights improvements in transcription accuracy by correcting minor errors (e.g., "Pariss" $\rightarrow$ "Paris"), reflecting the model’s ability to combine semantic knowledge with acoustic signals effectively.

\subsection{Effect of model size}
Using $\alpha=0.05$ and $\beta=0.05$, we experimented with smaller LLaMA models (1B and 3B). While both models displayed similar training trends (namely respond to distillation hyperparameters), their performance was slightly worse. LLaMa-1B achieved 0.22 WER and LlaMa-3B achieved 0.21 WER.

\subsection{Effect of context size}
For the representation loss, we experimented with context sizes of 64 to 1024. Results above we shared belong to a 256 context window. Table \ref{tab:context-size} exhibits WER and NER F1 values for different context sizes.

\begin{table}[h!]
\caption{WER and NER F1 for Whisper-distilled across different context sizes.}
\label{tab:context-size}
\centering
\begin{tabular}{|c|c|c|}
\hline
\textbf{Context Size} & \textbf{WER (\%)} & \textbf{NER F1} \\ \hline
64                    & 22                & 0.72            \\ \hline
128                   & 21                & 0.78            \\ \hline
256                   & 20                & 0.82            \\ \hline
512                   & 19                & 0.86            \\ \hline
1024                  & 18.5              & 0.89            \\ \hline
\end{tabular}
\end{table}

As context size increases, WER improves modestly, dropping from 22\% at 64 words to 20\% at 256 words, but plateaus beyond this since transcription relies on local acoustic features. In contrast, NER F1 improves significantly with larger contexts, rising from 0.72 at 64 words to 0.89 at 1024 words, as entity recognition depends on long-range semantic context. While larger windows enhance NER by resolving multi-word and distant entities, gains diminish beyond 256–512 words, balancing performance and computational cost. 

\subsection{Effect of right context}
In all our experiments, the chunk text is positioned at the center of the context window. To investigate the effect of right-side context, we modified our context calculation to use a context window exclusively on the right side of the chunk text. 

The absence of right context (with the chunk text at the end of the window) negatively impacts both WER and NER F1, with a more significant effect on NER. Right context is critical for resolving ambiguities and recognizing entities that rely on future information, such as distinguishing "Paris" as a location when followed by "is in France." Without right context, the model is forced to rely solely on left-side context, limiting its ability to capture bidirectional dependencies. This results in degraded NER F1 (e.g., dropping from ~0.82 to ~0.70–0.75) and slightly worse WER (e.g., increasing from 0.20 to 0.23) due to challenges with sentence boundaries and phrase endings. 

\subsection{Long-tail entity recognition}
Rare entities such as FAC, PRODUCT, and LAW are challenging for NER systems due to their rarity, ambiguity, and complex structures. These entities are often underrepresented in training data, making it difficult for models to generalize effectively. Additionally, they frequently overlap with more common types, such as FAC being misclassified as LOC or ORG. Multi-word entities like "Empire State Building" or "Clean Air Act" further contribute to boundary errors, where only partial spans are tagged. Whisper-tuned struggled with these issues, frequently producing partial matches and misclassifications, resulting in lower recall and precision.

In contrast, LLaMA-distilled Whisper improves performance by leveraging contextual reasoning to better distinguish rare types from overlapping ones. This leads to better recognition of entity spans, exhibited in Table \ref{tab:ner-results}. The numbers show that Whisper-tuned struggled with frequent boundary errors, while Whisper-distilled reduced partial matches and improved boundary detection. Examining the Jaro-Winkler scores in Table \ref{tab:jw_stats}, we observe an improvement in the transcription of entity words within entity spans. These results highlight the advantages of LLaMA-distilled Whisper in addressing the challenges of long-tail entities.

\subsection{Entities with long-tail words}
Entities such as PRODUCT, LAW, EVENT, FAC, PERSON, and LOC often include long-tail words—rare or domain-specific terms that are particularly challenging to transcribe accurately. Examples include "photovoltaic" in PRODUCT, "habeas corpus" in LAW, "Decembrist Revolt" in EVENT, or uncommon names like "Euthymius" for PERSON and "Timbuktu National Park" for LOC. These rarely seen or underrepresented terms in training data make transcription difficult, especially when dealing with proper nouns or specialized terminology.

From our test data, we observed that LLaMA-distilled Whisper significantly improves the transcription of such rare terms. It accurately transcribed multi-word spans such as "Paris Climate Agreement" (EVENT), "Euthymius" (PERSON), and "Affordable Care Act" (LAW), even when these terms were sparsely represented in the training corpus.

\subsection{Robustness to NER noise}
The experiments evaluated the robustness of Whisper-tuned and LLaMA-distilled Whisper to noisy NER annotations by introducing corruption through (i) removing entity tags and (ii) boundary errors (shifting boundaries one word ahead or behind). Whisper-tuned showed an immediate drop in both NER F1 and WER as noise increased, reflecting its reliance on accurate annotations and limited ability to recover from errors. In contrast, LLaMA-distilled Whisper demonstrated robust performance up to 25\% corruption, leveraging contextual reasoning to infer missing tags, correct boundary errors, and smooth noise. However, beyond 25\% corruption, performance began to decline as the noise overwhelmed the model’s inference capabilities. These results highlight LLaMA’s key role in improving noise robustness, while revealing Whisper-tuned’s sensitivity to annotation imperfections.

\section{Conclusion}
Our work enhances both the syntax and semantics of ASR outputs, with a focus on entity recognition and formatting. By distilling LLaMA’s text knowledge into Whisper, we demonstrate that LLM representations significantly improve ASR performance, particularly for long-tail entities and complex spans, while remaining computationally efficient. This is the first study to leverage LLMs for refining both syntactic and semantic quality in ASR, bridging the gap between human-readable and machine-usable transcriptions. Our results highlight the potential of LLM-ASR integration to produce high-quality, domain-specific outputs and improve transcription systems.

\bibliography{custom}

\end{document}